\documentclass{article}

\usepackage{hyperref}
\usepackage{microtype}
\usepackage{graphicx}
\usepackage{booktabs} %
\usepackage[utf8]{inputenc} %
\usepackage[T1]{fontenc}    %
\usepackage{hyperref}       %
\usepackage{url}            %
\usepackage{booktabs}       %
\usepackage{amsfonts}       %
\usepackage{nicefrac}       %
\usepackage{microtype}      %

\usepackage{siunitx}

\usepackage{amsmath}
\DeclareMathOperator{\enc}{Enc}
\DeclareMathOperator{\dec}{Dec}

\usepackage{amsthm}
\usepackage{graphicx}
\usepackage{xcolor}
\usepackage{comment}
\usepackage{subcaption}
\usepackage{multirow}
\usepackage{caption}
\usepackage{sidecap}
\usepackage{wrapfig}
\usepackage{xfrac}
\newtheorem{remark}{Remark}

\newcommand{\ci}{Coded-InvNet}
\newcommand{\ir}{i-ResNet}
\newcommand{\irev}{i-RevNet}
\newcommand{\parm}{ParM}
\newcommand{\lac}{Learning-a-code}
\newcommand{\dm}{MNIST}
\newcommand{\df}{Fashion-MNIST}
\newcommand{\dc}{CIFAR10}
\newcommand{\ptp}{Pix2Pix}
\newcommand{\mmix}{Manifold Mixup}

\newcommand{\app}{the supplementary material}

\usepackage[accepted]{icml2021}

\icmltitlerunning{\ci{} for Resilient Prediction Serving Systems}

\begin{document}

\twocolumn[
\icmltitle{\ci{} for Resilient Prediction Serving Systems}
\icmlsetsymbol{equal}{*}

\begin{icmlauthorlist}
\icmlauthor{Tuan~Dinh}{to}
\icmlauthor{Kangwook~Lee}{goo}
\end{icmlauthorlist}

\icmlaffiliation{to}{Department of Computer Sciences, 
  University of Wisconsin-Madison, 
  Madison, USA}
\icmlaffiliation{goo}{Department of Electrical and Computer Engineering, 
   University of Wisconsin-Madison, 
   Madison, USA}

\icmlcorrespondingauthor{Tuan~Dinh}{tuan.dinh@wisc.edu}
\icmlcorrespondingauthor{Kangwook~Lee}{kangwook.lee@wisc.edu}

\icmlkeywords{Distributed System, Coded Computation, GANs, Invertible Networks}

\vskip 0.3in
]
\printAffiliationsAndNotice{} %

\begin{abstract}
Inspired by a new coded computation algorithm for invertible functions, we propose \ci{}, a new approach to design resilient prediction serving systems that can gracefully handle stragglers or node failures.
\ci{} leverages recent findings in the deep learning literature such as invertible neural networks, \mmix{}, and domain translation algorithms, identifying interesting research directions that span across machine learning and systems.
Our experimental results show that \ci{} can outperform existing approaches, especially when the compute resource overhead is as low as $10\%$.
For instance, without knowing which of ten workers is going to fail, our algorithm can design a backup task to correctly recover the missing prediction with {an accuracy} of $85.9\%$, significantly outperforming the previous SOTA {by $32.5\%$.}%
\end{abstract}
\section{Introduction}
Prediction serving systems (PSSs) are systems that host pre-trained machine learning models for inference queries such as search, translation, and ranking ones.
PSSs usually consist of a large number of parallel/distributed compute workers to process a large batch of input queries in parallel. 
As the system scales, maintaining low response time (latency) becomes more challenging.
This is because the slowdown or failure of a single node can slow down the entire query processing time~\citep{dean2013tail,narra2019distributed}.
Multiple approaches for latency reduction have been proposed, including detect-and-relaunch~\citep{zaharia2008improving}, query replication~\citep{suresh2015c3}, and approximate inference~\citep{han2017clap}.

\emph{Coded computation}, by introducing backup tasks in a coded form, improves system resilience with lower overhead compared to other methods~\citep{lee2017speeding}. 
Recently,~\citep{kosaian2019parity} showed that a coded computation-based technique, called Parity Models (\parm{}s), can significantly reduce the latency of PSSs.
\parm{} consists of a tuple of encoder/parity model/decoder: the encoder aggregates multiple queries into a coded query, the parity model computes an inference task on it, and the decoder uses the encoded query together with available task results to reconstruct missing task results.
However, experimental results of \parm{} were limited to small-scale PSSs (2--4 workers), and it is not clear whether the proposed method is applicable to large-scale PSSs.
Also, small-scale regimes correspond to high resource overhead regimes, as one backup worker is required for every two to four workers.
We empirically show that \parm{} does not scale well. 
For instance, when there is only one backup worker for ten main workers, \parm{} achieves approximately $19\%$ accuracy on the \dc{} image classification when one of the main workers is not available.

We propose \ci{}, a new coded computation-based framework to design scalable resilient prediction serving systems.
Inspired by an efficient coded computation scheme for invertible functions, \ci{} designs the inference function with a computationally-heavy but invertible module followed by a computationally-light module.
To design the invertible module, we take advantage of recent developments in invertible neural networks~\citep{behrmann2018invertible,song2019mintnet,jacobsen2018revnet}.
Then, by making use of GAN-based paired domain translation algorithms~\citep{isola2017image},
\ci{} trains a light-weight encoding function so that one can efficiently generate encoded queries from input queries without incurring high encoding overhead.
To further improve the classification accuracy in the event of failures, \ci{} also leverages mixup algorithms~\citep{zhang2017mixup,verma2018manifold}.

We evaluate the efficacy of \ci{} in image classification and multitask learning setting. 
Experimental results show that \ci{} can scale much beyond the scales at which existing algorithms operate.
More specifically, \ci{} achieves higher reconstruction accuracy compared to \parm{}, and the accuracy gap increases remarkably as the system scales. 
For instance, \ci{} can achieve reconstruction accuracy of $85.9\%$ with $10$ workers, while that of \parm{} is $53.4\%$. 
We also evaluate end-to-end latencies on an AWS EC2 cluster, showing that \ci{}'s computing overhead is negligible.

\subsection{Key Idea of \ci{}}\label{sec:cc_for_invertiblefcn}
The key idea of coded computation is best illustrated when the target function is linear~\citep{lee2017speeding}. 
For illustration purposes, consider two inputs $x_1$ and $x_2$ and three parallel workers.
The goal here is to assign computation tasks to the workers so that one can obtain $f(x_1)$ and $f(x_2)$ even in the presence of slowdown or failure of a node.
Coded computation assigns $f(x_1)$ to the first worker and $f(x_2)$ to the second worker.
For the third worker, it first combines two input queries to get an encoded query $\frac{x_1+x_2}{2}$.
It then assigns $f\left(\frac{x_1+x_2}{2}\right)$ to the third worker. 
By the linearity of the target function $f(\cdot)$, we have $f\left(\frac{x_1+x_2}{2}\right) = \frac{f(x_1)+f(x_2)}{2}$.

These three tasks can be viewed as three linearly independent weighted sums of $f(x_1)$ and $f(x_2)$, i.e., 
\begin{align}
\begin{bmatrix}
f(x_1)\\
f(x_2)\\
\frac{f(x_1) + f(x_2)}{2}
\end{bmatrix}
= 
\begin{bmatrix}
1 & 0\\
0 & 1\\
\frac{1}{2} & \frac{1}{2}
\end{bmatrix}
\begin{bmatrix}
f(x_1)\\
f(x_2)
\end{bmatrix}.\label{eq:1}
\end{align}

Observe that \emph{any} two rows of the coefficient matrix in the RHS of \eqref{eq:1} is an invertible matrix.
That is, as long as any two of the three computation results are available, the decoder $\dec{(\cdot, \cdot)}$ can decode the computation results to recover $f(x_1)$ and $f(x_2)$. 
For instance, consider a situation where the second computation result $f(x_2)$ is missing. 
Then, 
\begin{align}
\begin{bmatrix}
f(x_1)\\
\frac{f(x_1) + f(x_2)}{2}
\end{bmatrix}
= 
\begin{bmatrix}
1 & 0\\
\frac{1}{2} & \frac{1}{2}
\end{bmatrix}
\begin{bmatrix}
f(x_1)\\
f(x_2)
\end{bmatrix}.\label{eq:2}
\end{align}
Since the coefficient matrix is full rank, the decoder can simply multiply the inverse of the coefficient matrix to recover $f(x_1)$ and $f(x_2)$.

\begin{figure}[t]
	\centering
	\includegraphics[width=0.98\linewidth]{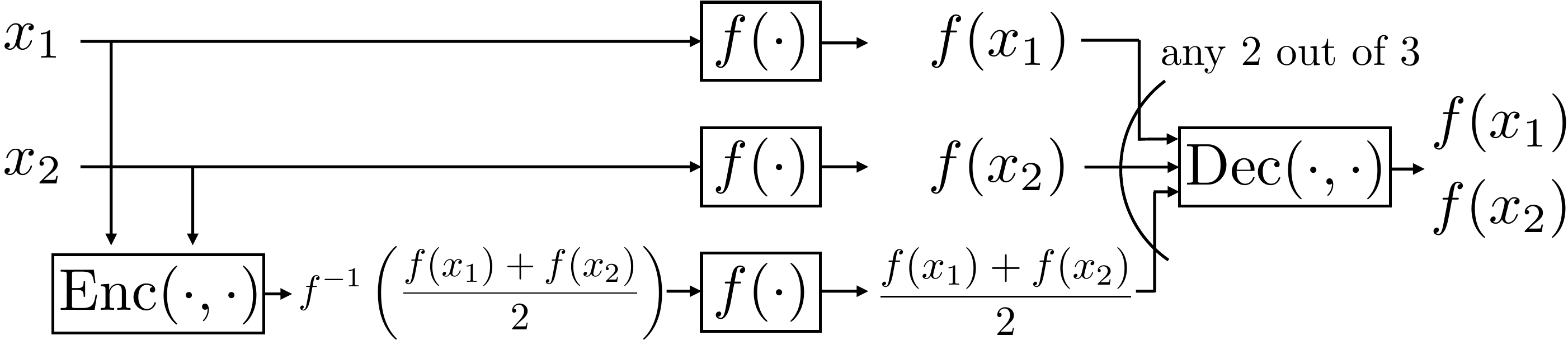}
    \caption{\textit{\textbf{Coded computation for invertible functions.} Consider an invertible function $f(\cdot)$ and two inputs $x_1$ and $x_2$. The optimal encoding function $\enc{(\cdot, \cdot)}$ takes the two inputs and generates an encoded input $f^{-1}\left(\frac{f(x_1) + f(x_2)}{2}\right)$. By applying function $f(\cdot)$ to the original inputs and the encoded input, one can obtain linearly independent weighted sum of $f(x_i)$'s. Thus, a simple decoder can always take \emph{any} 2 out of 3 computation results to decode $f(x_1)$ and $f(x_2)$ by solving a system of linear equations. For instance, if $f(x_2)$ is missing, one can multiply $\frac{f(x_1) + f(x_2)}{2}$ by $2$ and then subtract $f(x_1)$ to recover $f(x_2)$. Our framework \ci{} extends this idea to design resilient prediction serving systems.}}	
    \label{fig:code_invnet}
    \vspace{-0.5cm}
\end{figure}

\ci{} is based on a simple observation that this coded computation framework is applicable to the family of invertible functions.
Note that the target function is assumed to be linear or polynomial in the input in the existing works~\citep{lee2017speeding,pmlr-v89-yu19b}.
Shown in Fig.~\ref{fig:code_invnet} is the visual illustration of the coded computation algorithm for an invertible function. 
Consider the following encoding function $\enc{(x_1, x_2)}= f^{-1}\left(\frac{f(x_1) + f(x_2)}{2}\right)$.
By applying the target function to $x_1$, $x_2$, and $f^{-1}\left(\frac{f(x_1) + f(x_2)}{2}\right)$, we obtain $f(x_1)$, $f(x_2)$, and $\frac{f(x_1) + f(x_2)}{2}$.
Therefore, as shown in the previous example, the decoder can always recover $f(x_1)$ and $f(x_2)$ with \emph{any} two out of three task results.\footnote{{We provide a concrete example on synthesis dataset to complete this illustration in \app{}.}}

While this example demonstrates an efficient coded computation for invertible functions, this scheme `as it is' cannot be used in practice.
This is because the encoding process involves multiple evaluations of $f(\cdot)$ and $f^{-1}(\cdot)$, resulting in a significant encoding overhead.

However, in machine learning applications, one can resolve this issue for the following reasons.
First, $f(\cdot)$ does not need to be exactly computed, and it is sufficient to compute $f(\cdot)$ up to some approximation error.
For instance, if $f(\cdot)$ is the feature extractor, then a small approximation error may not alter results of the following downstream tasks such as classification, segmentation, etc.
This immediately implies that one can safely replace the computationally expensive encoding function with an approximate encoding function.
Second, $f(\cdot)$ is also a design choice in machine learning applications. 
For instance, if $f(\cdot)$ is a feature extractor used by a downstream classification task, one can choose any $f(\cdot)$ as long as the classification performance is maintained.

Inspired by these observations, \ci{} replaces the ideal encoding function with an approximate function that is parameterized by a light-weight neural network. 
Then, it \emph{learns} the encoding function $\enc{(\cdot)}$ together with the target function $f(\cdot)$.
By carefully designing the encoding function's architecture, one can obtain low enough encoding approximation error while maintaining the encoding overhead negligible.

\section{Related Works}

\textit{Straggler mitigation in prediction serving systems} has been studied in various approaches, such as detect-and-relaunch~\citep{zaharia2008improving}, replication~\citep{dean2013tail}, approximate computing~\citep{goiri2015approxhadoop}, and coded computing~\citep{lee2017speeding}.
Among these, coded computation requires a minimal amount of compute resource overhead, making it promising with many methods to provide resiliency to slowdowns and failures in machine learning PSS~\citep{lee2017speeding,li2016unified,kosaian2018learning,kosaian2019parity,narra2019distributed}.
\lac{}~\citep{kosaian2018learning,Dhakal_2019} learns a neural network-based encoder/decoder pair to transform erasure-coded queries into a form that decoders can reconstruct unavailable predictions.
However, this approach suffers from high encoding and decoding overhead.
\parm{}~\citep{kosaian2019parity} overcomes this limitation by 
learning a new inference function applied to encoded queries, which they call parity models. 
\citep{narra2019distributed} proposes a novel convolutional neural network for multi-image classification on a collage image in one shot, thus reducing cost redundancy. 
This design is specific for image classification while \ci{} is applicable to more downstream tasks.

\textit{Invertible Neural Networks (INNs)} are NNs that can be inverted from the input up to the projection, or final classes~\citep{jacobsen2018revnet, behrmann2018invertible, song2019mintnet}.
\irev{}s~\citep{jacobsen2018revnet} define an INN family based on a succession of homeomorphic layers.
\citet{behrmann2018invertible} builds \ir{} on top of residual networks~\citep{he2016deep}, showing that one can invert a residual block with an exponential convergence rate via fixed-point iteration if Lipschitz constants of nonlinear parts are strictly less than 1. 
The authors bound each layer's Lipshitz constant by normalizing the weight matrix by its spectral norm.
We adopt \ir{} for our INN's architecture.

\textit{\mmix{}}~\citep{verma2018manifold}  extends Mixup~\citep{zhang2017mixup} to augment classification models with virtual data points which are random convex combinations of data points in the hidden space.
This simple scheme effectively improves the margin in the hidden space, helps flatten the class-conditional representation, and reduces the number of directions by a significant variance. 
\ci{} uses the classifier augmented with \mmix{} to make it more robust to encoding approximation errors.

\textit{\ptp{}}~\citep{isola2017image} is an image-to-image translation model that converts a set of images into a target image. 
\ptp{} trains a conditional GAN~\citep{mirza2014conditional} on pairs of input and target images, using a combination of GAN and $L_1$ losses.
\ptp{} uses Patch-GAN~\citep{isola2017image} and U-Net~\citep{ronneberger2015u} for the discriminator and generator architectures, respectively.
We use \ptp{} model for our encoder.

\section{\ci{}}

\subsection{Setting}
Given a task needed to be deployed, we train an invertible network followed by a neural network that has significantly fewer layers and parameters compared to the first part.
While this specific architectural choice may seem restrictive, it indeed supports a large range of applications. 
First of all, for classification tasks, state-of-the-art invertible NNs nearly match the performance of non-invertible NNs, and the gap is quickly closing.
For instance, MintNet~\citep{song2019mintnet} achieves only 1.4\% lower accuracy on \dc{} than ResNet.
Moreover, even non-invertible NNs are pseudo-invertible when trained with adversarial training~\citep{engstrom2019adversarial}, so our framework can serve any robust classification model.
Though we limit our focus on the classification in this work, \ci{} can be immediately extended to support generative models that are based on invertible architectures, such as flow-based models~\citep{kingma2018glow}, {or invertible Transformer-based models~\citep{kim2020lipschitz}}.

\begin{figure*}
    \centering
    \includegraphics[width=0.85\textwidth]{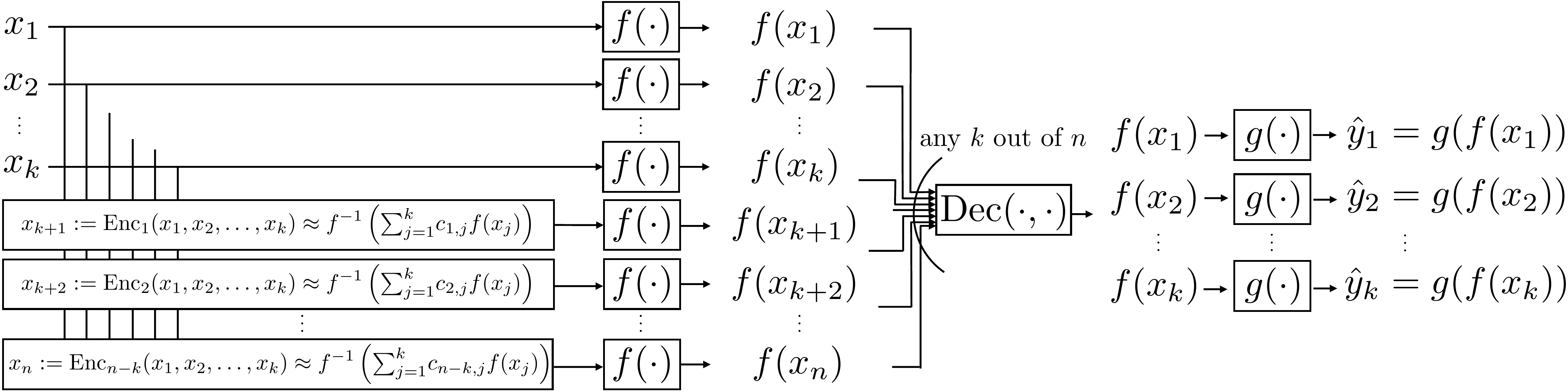}
    \caption{\textbf{Prediction serving system architecture of \ci{}.} 
    \textit{Given $k$ inputs $x_1, x_2, \ldots, x_k$, the goal of the predicting serving system is to compute $\hat{y}(x_i) := g(f(x_i))$ for all $1 \leq i \leq k$ in the presence of stragglers or failures, where $f(\cdot)$ is an invertible function. To this end, \ci{} first generates $n-k$ encoded inputs $x_{k+1}, \ldots, x_{n}$ by applying $n-k$ distinct encoding functions to $x_i$'s. It then assigns the task of computing $f(x_i)$ to each of $n$ parallel workers. By leveraging the coded computation algorithm for invertible functions described in Sec.~\ref{sec:cc_for_invertiblefcn}, one can approximately decode $f(x_1), f(x_2), \ldots, f(x_k)$ as soon as any $k$ out of $n$ tasks are completed. Note that the approximation errors occur since we use approximate encoding functions. Once $f(x_1), f(x_2), \ldots, f(x_k)$ are computed, the front end applies $g(\cdot)$ to each of them and returns the query results.}}
    \label{fig:model}
    \vspace{-0.1in}
\end{figure*}
Shown in Fig.~\ref{fig:model} is the PSS architecture of \ci{}.
Here, the overall inference network is denoted by $\hat{y}(x) := g(f(x))$, where $f(\cdot)$ is the invertible module of the neural network and $g(\cdot)$ is the second module of the network.
The goal is to compute $\hat{y}_1 = g(f(x_1)), \hat{y}_2 = g(f(x_2)), \ldots, \hat{y}_k = g(f(x_k))$ in parallel with $n$ ($n\geq k$) parallel workers in the presence of stragglers or failures. 
At the time the query arrives, the front end of the PSS applies $n-k$ encoding functions to obtain $n-k$ encoded queries.
The $i$-th encoding function is denoted by $\enc_{i}{(\cdot)}$, and the $i$-th encoded query is denoted by $x_{k+i}$ for $1 \leq i \leq n-k$.
That is, $x_1, x_2, \ldots, x_k$ denote the $k$ inputs while $x_{k+1}, x_{k+2}, \ldots, x_n$ denote the $n-k$ encoded inputs.

Similar to the example illustrated in Sec.~\ref{sec:cc_for_invertiblefcn}, we first identify the design of $n-k$ ideal encoding functions as follows.
An ideal encoded input is such that one can get a linear combination of $f(x_i)$'s by applying $f(\cdot)$ to it.
Let $f(x_{k+i}) = \sum_{j=1}^{k}{c_{i,j}f(x_j)}$ for $1 \leq i \leq n-k$.
Then,
\vspace{-0.5em}
\small
\begin{align}
\setlength\arraycolsep{1pt}
    \!\!\!\!\!\!\begin{bmatrix}
    f(x_1)\\f(x_2)\\\vdots\\f(x_k)\\f(x_{k+1})\\\vdots\\f(x_{n})\end{bmatrix}
    = \begin{bmatrix}
    1 & 0 & \cdots & 0\\
    0 & 1 & \cdots & 0\\
    \vdots & \vdots & \ddots & \vdots\\
    0 & 0 & \cdots & 1\\
    c_{1,1} & c_{1,2} & \cdots & c_{1,k}\\
    \vdots & \vdots & \ddots & \vdots\\
    c_{n-k,1} & c_{n-k,2} & \cdots & c_{n-k,k}
    \end{bmatrix}
    \begin{bmatrix}
    f(x_1)\\f(x_2)\\\vdots\\f(x_k)
    \end{bmatrix}.\label{eq:eq3}
\end{align}
\normalsize
To guarantee the decodability when any $k$ tasks complete, any $k$ rows of the coefficient matrix in the RHS of \eqref{eq:eq3} must be full rank.
For instance, when $n = k +1$, i.e., there is only one additional worker, one can satisfy this property by setting $c_{1,j} = \frac{1}{k}$ for all $j$. 
Also, when $c_{i,j}$ is chosen from the i.i.d. Gaussian distribution, one can show that the coefficient matrix satisfies the property with probability $1$. 
Referring to the readers to \citep{Bodmann2013} for various ways of choosing $c_{i,j}$'s, we will assume that $c_{i,j}$'s are chosen such that the desired property holds.

\begin{remark}
When choosing $c_{i,j}$'s, in addition to the decodability condition, one should also consider whether $f(x_{k+i}) = \sum_{j=1}^{k}{c_{i,j}f(x_j)}$ lies close enough to the original manifolds in the embedding space.
This is because the invertibility of $f^{-1}$ may not hold for embedding vectors that are too far away from the original manifolds. 
Moreover, if this additional condition holds, $f^{-1}(f(x_{k+i}))$ might share some semantic features with the original data.
This essentially reduces the gap between the domain where original inputs belong and the domain where encoded inputs belong, allowing for efficient learning of encoding functions with domain translation techniques.
\end{remark} 

\subsection{Approximate Encoding Functions}
Recall that ideal encoding functions cannot be used in practice due to their large computation overhead.
Thus, we use a neural network (NN) to approximate ideal encoding functions.
The universal approximation theorem asserts that ideal encoding functions can be well approximated by some NNs~\citep{cybenko1989approximation}. 
By limiting the number of layers and parameters of the NN, we can control the computation overhead of the NN-based approximate encoding functions.

More specifically, for each $i$, $1\leq i \leq n-k$, we want to train a neural network $\enc_{i}{(\cdot)}$ such that $\enc_{i}{(\cdot)} \approx f^{-1}\left(\sum_{j=1}^{k}{c_{i,j}f(x_j)}\right)$.
Note that if $x \in \mathcal{X}$, then $\enc_{i}{(\cdot)}: \mathcal{X}^k \rightarrow \mathcal{X}$, i.e., the encoding function takes $k$ inputs from domain $\mathcal{X}$ to generate one output in $\mathcal{X}$. 
Thus, to learn the $i$-th encoding function, we need to collect the following train set:
$
\left\{(x_1, x_2, \ldots, x_k), f^{-1}\left(\textstyle{\sum_{j=1}^{k}}{c_{i,j}f(x_j)}\right)\right\}_{}^{}
$.
Here, the input $(x_1, x_2, \ldots, x_k)$ is a tuple of $k$ randomly chosen inputs from the train set.
Given the input tuple, the `label' $f^{-1}\left(\textstyle{\sum_{j=1}^{k}}{c_{i,j}f(x_j)}\right)$ can be always computed by applying $f(\cdot)$ and $f^{-1}(\cdot)$. 
Once the train set for an encoding function is collected, one can train the encoding function such that $\enc_{i}{(\cdot)} \approx f^{-1}\left(\sum_{j=1}^{k}{c_{i,j}f(x_j)}\right)$ holds on the collected train set.

One can also make use of explicit semantic loss when training encoding functions.
For illustration purposes, assume that $n = k + 1$ and $c_{1,j} = \frac{1}{k}$ for all $j$.
When $f(x_a)$ is missing for some $1\leq a \leq k$, and all the other task results are available, one can decode $f(x_i)$ as follows:
$\widehat{f(x_a)} = k f(x_{k+1}) - \sum_{i=1,\ldots,k, i\neq a}^{k}f(x_i)$.
And the inference result will be $g\left(\widehat{f(x_a)}\right)$.
By comparing this inference result with the correct result $g\left(f(x_a)\right)$, one can explicitly capture the semantic difference between them. 
For instance, if the target task is classification, one can compare the logit values of these outputs and apply the distillation loss~\citep{hinton2015distilling}. Note that one can also `co-train' these encoding functions together with the target function $g\circ f$.
 
\begin{remark}
    One can further reduce the {inference overhead of the encoder} by applying neural network compression techniques such as pruning, knowledge distillation, vector quantization, etc.
Such techniques allow for larger encoder architectures, reducing the encoding error.
\end{remark}
\begin{remark}
One can deploy $\enc{(\cdot, \cdot)}$ at the front-end server or at the backup workers. 
The former option increases the computational load of the front-end server while the latter increases the load of backup workers. 
Moreover, the latter incurs extra communication cost between the front-end and the workers. 
Thus, one should make a proper choice considering the communication/computation tradeoff.
\end{remark}

\subsection{Minimizing Encoding Error Propagation}\label{sec:mixup}
Once we introduce approximate encoding functions into this framework, $f(x_{k+i})$, for $1 \leq i \leq n-k$, will not exactly match the desired linear combination $\sum_{j=1}^{k}{c_{i,j}f(x_j)}$.
That is, with approximate encoding, we only have $x_{k+i} \approx f^{-1}\left(\sum_{j=1}^{k}{c_{i,j}f(x_j)}\right)$, so $f(x_{k+i}) = \sum_{j=1}^{k}{c_{i,j}f(x_j)} + \varepsilon$, where $\varepsilon$ is the approximation error in $f$, which depends on $(x_1,x_2,\ldots,x_k)$. 

To see how this approximation error affects the inference quality, assume that $n = k + 1$ and consider a scenario where $f(x_1)$ is missing, and $f(x_2), f(x_3), \ldots, f(x_k), f(x_{k+1})$ are available.
By assuming $c_{1,j} = \frac{1}{k}$, one can decode $f(x_1)$ as follows:
$\widehat{f(x_1)} = k f(x_{k+1}) - \sum_{i=2}^{k}f(x_i)$.
Since $f(x_{k+1}) = \sum_{j=1}^{k}{c_{1,j}f(x_j)} + \varepsilon$, we have $\widehat{f(x_1)} = f(x_{k+1}) + k\varepsilon$.
Then, $\hat{y}_1 = g(f(x_{k+1}) + k\varepsilon)$. 
Therefore, the recovered inference result for a missing computation task will be incorrect if $\hat{y}_1 = g(f(x_{k+1}) + k\varepsilon) \neq g(f(x_{k+1}))$.

To prevent this, one must first make sure that the encoding functions are well trained such that $\enc_{i}{(\cdot)} \approx f^{-1}\left(\sum_{j=1}^{k}{c_{i,j}f(x_j)}\right)$ holds.
That way, one can expect that the magnitude of $\varepsilon$ is small enough.

However, even when encoding error $\varepsilon$ is small, if the second part of the inference module $g(\cdot)$ is highly sensitive to small perturbations, inference results could still be incorrect.
Indeed, \citet{verma2018manifold} shows that standard deep neural networks are highly sensitive to small noises injected in the embedding space.
This implies that a small encoding error, which necessarily arises during the approximate encoding procedure, can distort the overall inference result. 

To resolve this issue, we leverage the recent \mmix{}~\citep{verma2018manifold} regularization technique, which smoothens class-conditional embedding spaces, making inference outputs more robust to noises injected in the embedding space. 
Inspired by this, \ci{} trains $f(\cdot)$ and $g(\cdot)$ with \mmix{} to ensure that 
$\hat{y}_1 = g(f(x_{k+1}) + k\varepsilon) = g(f(x_{k+1}))$ with high probability.

\subsection{Advantages of \ci{}}
\paragraph{Separation between Encoding and Inference}
By design, \ci{} maintains a clear separation between encoding and inference,\footnote{\lac{}~\citep{kosaian2018learning} also has this property.} which is highly advantageous when building a prediction serving system.
First, every worker does the same work, i.e., simply computing $f(\cdot)$, simplifying the system implementation and management.
On the other hand, \parm{} requires a highly heterogeneous system configuration: the first $k$ workers compute $f(\cdot)$, and the other $n-k$ workers compute $n-k$ \emph{distinct} parity models, say $f'_1(\cdot), \ldots, f'_{n-k}(\cdot)$.
Similarly, when the total number of workers $n$ and the total number of inputs $k$ change, \ci{} does not require any change to the worker configuration and simply needs to retrain the encoder.

Second, when optimizing inference time via various techniques such as model compression~\citep{zhang2019deep}, hardware optimization~\citep{marculescu2018hardware}, or network pruning~\citep{blalock2020state}, one just needs to focus on optimizing one inference function $f(\cdot)$. 
On the other hand, \parm{} needs to optimize $n-k+1$ models, incurring a significantly larger cost.

\paragraph{Applicability to Multi-task Serving}
A representation in NNs can serve for various downstream tasks~\citep{liu2015representation}.
For instance, well-trained image representation can be simultaneously useful for image classification, segmentation, and depth estimation~\citep{kendall2018multi}.
For a detailed overview of multi-task learning and the role of shared representation, we refer the readers to~\citep{ruder2017overview}.

We highlight that \ci{} is a promising solution when the prediction serving system is serving multiple inference tasks sharing the underlying representation.
Recall that \ci{} reconstructs the embedding (or representation) of missing inputs while existing approaches directly reconstruct missing inference results~\citep{kosaian2018learning,kosaian2019parity}.
Therefore, \ci{} does not need any extra efforts to support multiple tasks, say $m$ tasks.
The only difference is that we now have $g_i(\cdot)$ for $1\leq i \leq m$, where $g_i(\cdot)$ is specific to task $i$.
Note that this does not affect the training complexity of the encoder at all. 
On the other hand, \parm{} has to train $m$ parity models to achieve the same goal.
See Sec.~\ref{sec:exp_mtl} for experimental results where we demonstrate the applicability of \ci{} to multi-task settings.

\section{Experiments}
We show how we implement the components of the \ci{} framework and evaluate its performance.
In comparison with the baselines, we focus on the image classification task on popular datasets: \dm{}~\citep{deng2012mnist}, \df{}~\citep{xiao2017/online}, and  \dc{}~\citep{krizhevsky2009learning}. 
These are all $10$-way image classification tasks. 
{We also demonstrate the applicability of \ci{} on a large-scale ImageNet-based dataset~\cite{imagenette}, and on the multiple-failure setting.}

We report various performance metrics such as the classification accuracy when straggler/failure happens, encoding/decoding overhead, scalability, end-to-end latency, etc.

Since \ci{} recovers the full embedding of the missing input, it naturally fits with multi-task applications, i.e., one common embedding can be used for multiple downstream tasks.
We also show how one can apply \ci{} for such multi-task applications. 

We provide further additional experiment results and implementation details in \app{}.

\subsection{Architectures and Training Methods}\label{sec:exp_architectures}

While the \ci{} framework is applicable to any values of $n, k$, $n \geq k$, the most important case is when $n = k + 1$. 
This is of practical interest since this scheme's compute overhead is minimal ($\frac{100}{k}\%$) while still being robust against a single failure.
We will consider $k \in \{2, 4, 10\}$ and choose $c_{1,j} = \frac{1}{k}$ for all $j$.

\begin{figure}[t]
\vspace{-0.6em}
	\centering
	\includegraphics[width=.8\linewidth]{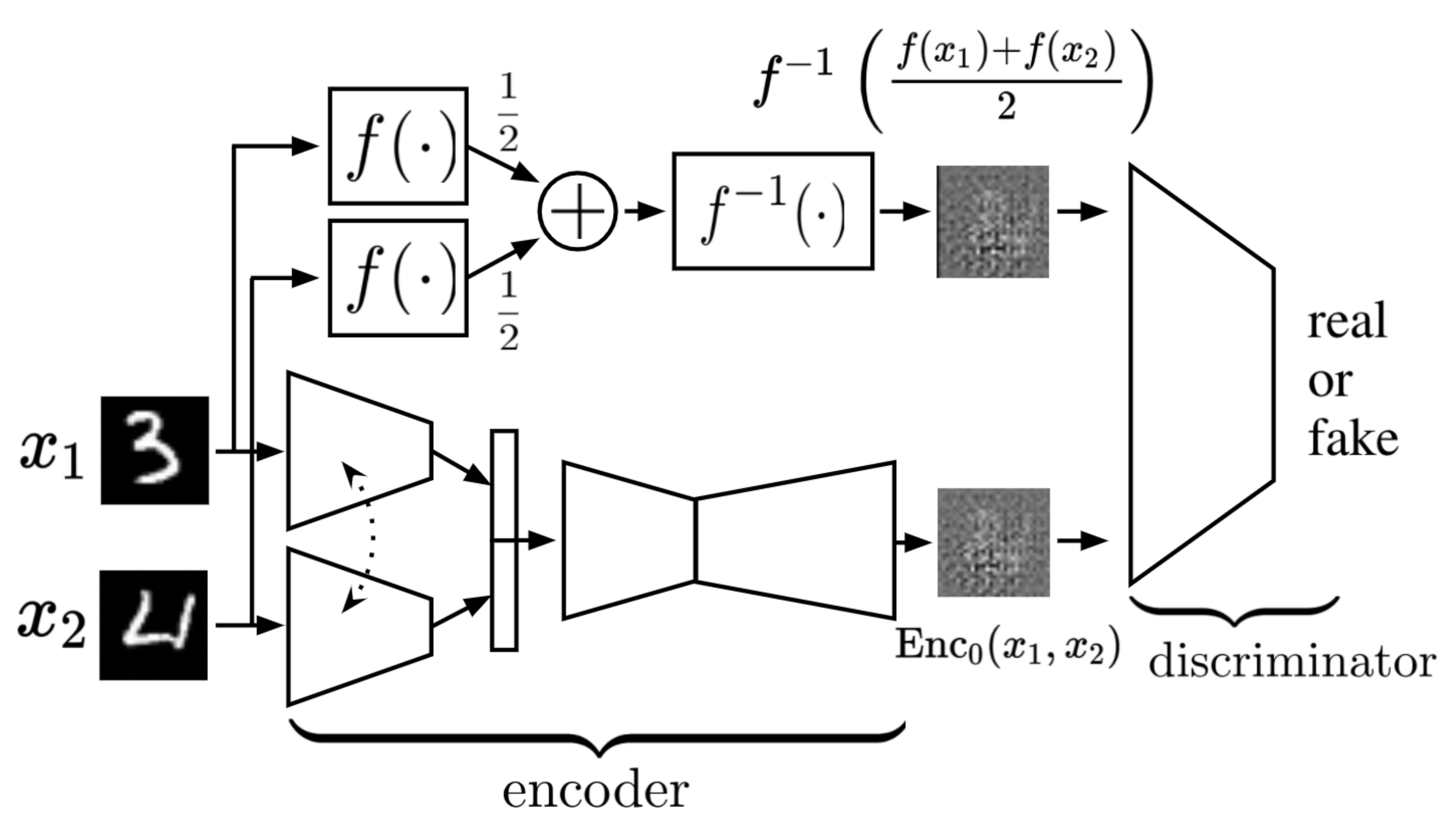}
	\vspace{-0.1in}
    \caption{\textit{\textbf{Encoder training.} Let $k=2$ and $c_{1,1}=c_{1,2}=\frac{1}{2}$. We first pick two random inputs $x_1$ and $x_2$. We first apply $f(\cdot)$ to each of them, compute the average, and then apply $f^{-1}(\cdot)$, obtaining the target. We then feed the two inputs to the encoder. We ensure the input symmetry of the encoder by individually processing them with a weight-shared network. The processing results are concatenated and then get further processed to obtain the encoded input. We then train the encoder's weights with the Pix2Pix loss.}}
    \label{fig:encoder}
    \vspace{-0.15in}
\end{figure}

\textit{Architecture}. We use \ir{} as the classification network $g \circ f$.\textcolor{blue}{\footnote{We report additional results with another invertible network \irev{}~\citep{jacobsen2018revnet} in \app{}.}} 
We mostly follow the recommended configurations in~\citep{behrmann2018invertible}, but we remove the injective padding module to improve the invertibility of off-manifold embedding vectors.
We use the \ptp~\citep{isola2017image} architecture for encoding functions in our experiments for all values of $k$. 
To avoid linear scaling, we design our encoder architecture such that only the complexity of the first few layers depends on $k$, and the rest of the architecture does not scale with $k$. 
See Fig.~\ref{fig:encoder} for the encoder architecture for $k=2$.
When $k=2$, two inputs $x_1$ and $x_2$ are first processed by the weight-shared network.
The two processing results are then concatenated and projected to a fixed-size hidden vector.
In general, our encoder processes $k$ inputs in parallel, and the concatenated output is projected to the same size hidden vector.
By limiting the size of this input processing part, we control how the encoder complexity scales as $k$ increases. 
See Sec.~\ref{sec:exp_encoder_overhead} for experimental results where we demonstrate that encoding overhead can be kept nearly constant for increasing values of $k$.

\textit{Training}. 
We train our \ir{} classifier with \mmix{} (with mixup coefficient 1).
Then, we use the trained classifier to generate a train set for the encoding network. 
In particular, we draw $k$ random inputs $(x_1, x_2, \ldots, x_k)$ and compute labels $f^{-1}\left(\textstyle{\sum_{j=1}^{k}}{c_{i,j}f(x_j)}\right)$ to obtain an input/output pair.
Here, since our $f(\cdot)$ is an \ir{} module, the inverse function does not have an explicit form, we compute the inverse function by solving fixed point equations.
We repeat this $50,000$ times to construct a sufficiently large training set.
We, then, train the encoder function using Least Square GAN loss~\citep{mao2017least} plus $100$ times $L_1$ loss.
We enforce permutation invariance of inputs by (1) sharing weights of the first hidden layers across all $k$ inputs and (2) using the average function after the first hidden layer. 
The encoder training procedure is summarized in Fig.~\ref{fig:encoder}.
{Training the encoder may take up to 7.5 hours (ResNet-301-based architecture with 150 epochs on a 48-GB RTX8000 GPU).
}
\subsection{Training Results}
\textbf{Classifier}
The trained classifiers achieve 99.2\%, 91.8\%, and 90.9\% on \dm{}, \df{}, and \dc{}, respectively. 
Note that removing the injection padding layer from the original \ir{}, as mentioned previously, results in slight drops in the classification accuracy.

\begin{figure*}[t]
    \centering
    \includegraphics[width=0.7\textwidth]{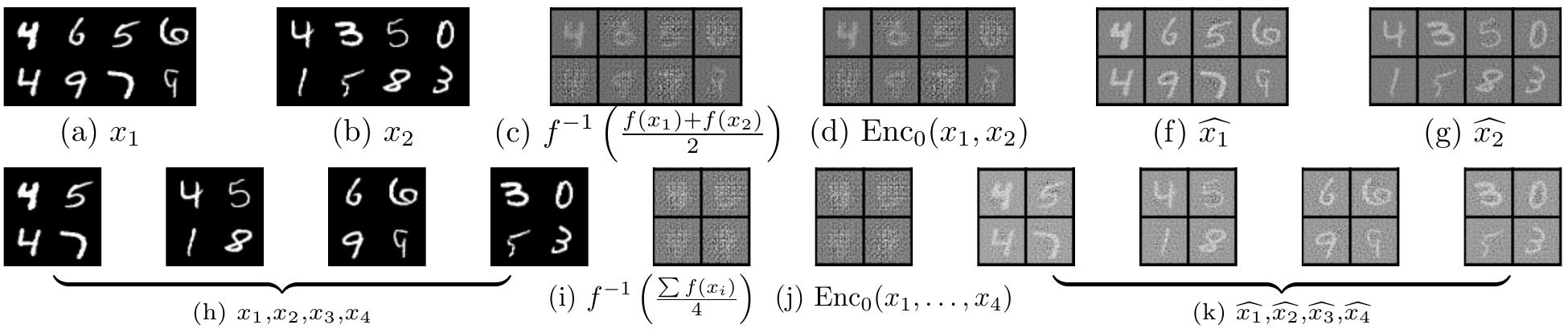}
    \vspace{-0.1in}
    \caption{\textbf{Qualitative performance evaluation of a trained encoder for MNIST ($k=2$ and $k=4$).} 
    \textit{
    We used $c_{1,j}=\tfrac{1}{k}$. 
    (a), (b): Randomly chosen $x_1$ and $x_2$.
    (c): The ideal encoding results, i.e., $f^{-1}\left((f(x_1) + f(x_2))/2\right)$.
    (d): The output of a trained encoder, i.e., $\enc_0{(x_1, x_2)}$. Note that (c) and (d) are visually indistinguishable.
    (f), (g): Note that $x_1 := f^{-1}\left(2f\left(f^{-1}\left((f(x_1) + f(x_2))/2\right)\right)-f(x_2)\right)$.
    By replacing $f^{-1}\left((f(x_1) + f(x_2))/2\right)$ with $\enc_0{(x_1, x_2)}$, we have $\widehat{x_1} :=  f^{-1}\left(2f(\textrm{Enc}_{0}{(x_1, x_2)})-f(x_2)\right)$. Similarly, one can also define $\widehat{x_2}$.
    Shown in (f) and (g) are $\widehat{x_1}$ and $\widehat{x_2}$.
    Observe that they successfully recover the missing inputs up to offsets.
    (h)--(k): We repeated the same experiment for $k=4$ and observed similar results.
    }
    }
    \label{fig:encoder_qualitative}
    \vspace{-0.4cm}
\end{figure*}

\textbf{Encoder} To see if the encoder training was successful, we observe the training curve (shown in Fig.~\ref{fig:encoder_curves} in \app{}). 
Here, we show the $L_1$ loss measured on the train/test sets of \dm{}, respectively.
The encoder was trained with $k=4$ and $c_{1,j} = \tfrac{1}{4}$.
The train $L_1$ loss keeps improving while the test $L_1$ loss saturates around epoch $20$. 
Shown in Fig.~\ref{fig:encoder_qualitative} is the qualitative performance evaluation of the trained encoder.
In Fig.~\ref{fig:encoder_qualitative}~(h), we show eight random samples of $(x_1,x_2,x_3,x_4)$ from the test set.
Fig.~\ref{fig:encoder_qualitative}~(i) visualizes ideal encoded inputs.
Recall that each of these ideal encoded inputs is computed by applying $f(\cdot)$ to each of $x_i$'s, obtaining a weighted sum, and then computing $f^{-1}(\cdot)$.
Since \ir{} does not have an explicit inverse function, $f^{-1}(\cdot)$ is computed by solving a fixed point equation via an interactive method.
In Fig.~\ref{fig:encoder_qualitative}~(j), we show that the encoded inputs obtained by our trained encoder are indistinguishable from ideal encoding results to naked eyes.

To further check the validity of the trained encoder, we perform a qualitative evaluation by observing the visual quality of input decoding. 
Note that with the ideal encoding input, one can perfectly recover missing inputs.
Assuming $n=k+1$ and $c_{1,j}=\tfrac{1}{k}$, let us define
\small
\begin{align}
&\widehat{f(x_1)} := kf\left(\textrm{Enc}_{0}{\left(x_1, x_2, \ldots, x_k\right)}\right)-\textstyle{\sum_{i=2}^{k}}f(x_k),\label{eq:decode}\\
&\!\!\!\!\!\!\!\widehat{x_1} := f^{-1}\left(kf\left(\textrm{Enc}_{0}{\left(x_1, x_2, \ldots, x_k\right)}\right)-\textstyle{\sum_{i=2}^{k}}f(x_k)\right)\!\!.
\vspace{-0.3in}
\end{align}
\normalsize
That is, $\widehat{f(x_1)}$ is the reconstruction of the missing function evaluation, and $\widehat{x_1}$ is the reconstruction of the corresponding input.
One can define $\widehat{f(x_i)}$ and $\widehat{x_i}$ for $2\leq i\leq k$ in a similar way.
If the encoding function is ideal, i.e., $\enc_{0}{\left(x_1, x_2, \ldots, x_k\right)} = f^{-1}\left(\textstyle{\sum_{j=1}^{k}}{f(x_j)}/k\right)$, then $\widehat{f(x_i)} = f(x_i)$ and hence $\widehat{x_i} = x_i$.
Therefore, we can indirectly evaluate the performance of an encoder by comparing $\widehat{x_i}$ and $x_i$.
See Fig.~\ref{fig:encoder_qualitative}~(c), (d) for the comparison results for $k=2$ and Fig.~\ref{fig:encoder_qualitative}~(i), (j) for $k=4$.
Observe that $\widehat{x_i}$ closely recovers $x_i$ in all tested cases up to some offsets, justifying the validity of the trained encoder.

\subsection{Degraded Mode Accuracy}
\paragraph{Comparison with Baselines}
We now evaluate the classification performance of \ci{} under the presence of stragglers or failures.
\emph{Normal accuracy} is simply the top-1 classification accuracy {of our classifier}, i.e., $\mathbb{E}_{x,y}\left[\hat{y} = y\right]$.
If $f(x_i)$ is available at the time of $k$ tasks complete, then $\widehat{f(x_i)} = f(x_i)$.
Therefore, the normal accuracy will capture the classification accuracy for such cases. 

We also define \emph{degraded mode accuracy}~\citep{kosaian2019parity}.
If any of $f(x_i)$ is not available at the time of $k$ task results are available, we first decode $\widehat{f(x_i)}$ using~\eqref{eq:decode} and then compute $\widehat{y_i} = g(\widehat{f(x_i)})$, the recovered inference result on $x_i$.
Due to the encoding error, $\widehat{y_i}$ could be different from $g(f(x_1))$.
The degraded accuracy is the accuracy measured with respect to this recovered inference result $g(\widehat{f(x_1)})$. 
We expect that the degraded accuracy is lower than the normal accuracy. 
More formally, it is defined as $\mathbb{E}_{x_1,y_1,x_2,x_3,\ldots,x_k}\left[g(\widehat{f(x_1)}) = y_1\right]$, where the expectation is over $k$ random inputs.

\begin{figure}[t]
    \centering
    \includegraphics[width=0.48\textwidth]{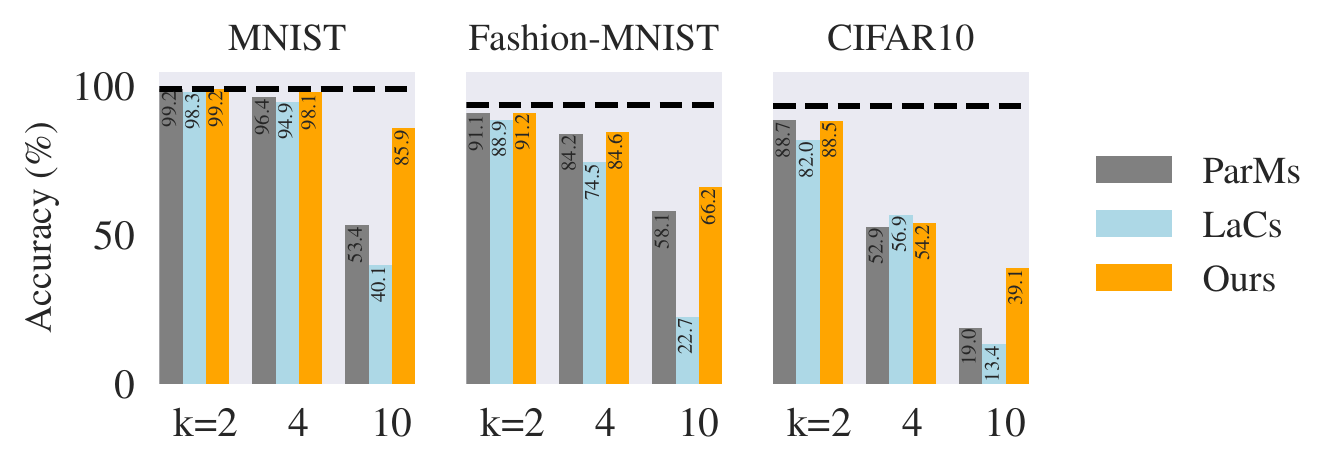}
    \vspace{-0.3in}
    \caption{\textbf{Degraded mode accuracy comparison.} 
    \textit{
    We compare \parm{}, \lac{} (LaC), and \ci{} (Ours) on different resource overhead $k = 2, 4, 10$.
    Dotted lines represent our normal accuracy $\mathbb{E}_{x,y}[\hat{y} = y]$, i.e., the classification accuracy when the inference does not involve any reconstruction. 
    Bars represent the degraded mode accuracy $\mathbb{E}_{x_1,y_1,x_2,x_3,\ldots,x_k,}[\hat{y} = y]$.
    Observe that all models achieve comparable performances when the resource overhead is high, e.g., $k=2$ ($50\%$) and $k = 4$ ($25\%$). 
    However, when the resource overhead is very low, i.e., $k=10$ ($10\%$), \ci{} significantly outperforms on every dataset.}
    }
    \label{fig:classificationacc}
    \vspace{-0.2in}
\end{figure}
We report the normal and degraded mode accuracy in Fig.~\ref{fig:classificationacc}. 
We compare our model with \parm{}~\citep{kosaian2019parity} built with the default summation encoder/subtraction decoder,\footnote{
\citet{kosaian2019parity} also develops a downsample-and-concatenation-based encoder for image classification, but this design is task-specific, and currently does not support $k=10$.}
and \lac{}~\citep{kosaian2018learning} built with convolutional encoder/fully-connected decoder.
{The dotted lines represent normal accuracies on each dataset, and the bars are for the degraded mode accuracies. }

When $k=2$ or $k=4$, all approaches obtain similar degraded accuracies on every dataset. 
When $k=10$, i.e., the system is larger and the resource overhead is much lower ($1/k = 10\%$), a large number of embedding vectors have to be packed into a single aggregated vector.
Also, the encoded input has more chance to be off the input manifold so is the corresponding embedding.
These factors make the encoder learning more challenging, indicating a trade-off between resource overhead and degraded accuracy, i.e., as the resource overhead decreases (as $k$ increases), the encoding and the degraded mode errors increase.
Indeed, in the original paper, the state-of-the-art approach \parm{} was tested only for $k\leq 4$, and it was not clear how well this approach would perform in such a low resource regime. 

As shown in Fig.~\ref{fig:classificationacc}, \ci{} significantly outperforms \parm{} and \lac{} in the low resource regime.
For instance, on \dm{}, \ci{} achieves $85.9\%$ degraded accuracy while \parm{} and \lac{} achieve $53.4\%, 40.1\%$ respectively. 
On \dc{}, the degraded mode accuracy of \ci{} is $39.1\%$, more than twice that of \parm{}, and three times that of \lac{}.

\textit{Take-away: \ci{} is highly resilient in the presence of stragglers or failures. It matches the performance of state of the art when the resource overhead is high enough ($\geq25\%$), and significantly outperforms them when the resource overhead is as low as $10\%$.}

\paragraph{{\ci{} on Large-scale ImageNet Data}}
\begin{figure}
    \centering
    \includegraphics[width=0.45\textwidth]{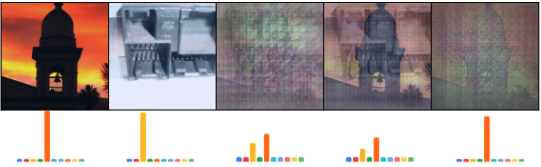}
    \vspace{-1em}
    \caption{\textbf{Reconstruction on Imagenette samples}. 
    \textit{Left to right: $x_1$, $x_2$, $f^{-1}((f(x_1)+f(x_2))/2)$, $\text{Enc}(x_1, x_2)$, and $\widehat{x_1}$. 
    Shown below them are the softmax scores of their prediction outputs $f(\cdot)$. 
    Note that $\arg\max f(\widehat{x_1}) = \arg\max f(x_1)$, \emph{i.e.}, the degraded classification is correct.}}
    \label{fig:imagenette}
    \vspace{-.15in}
\end{figure}
We show that our framework can be designed for more complex datasets.
In particular, we use Imagenette that is a 10-class subset of ImageNet. 
We replace \ir{} with an i-RevNet~\citep{jacobsen2018revnet} (96.2\% in normal accuracy). 
We achieve 85.5\% degraded accuracy for $(n,k)=(3,2)$ and 58.4\% for $(n,k)=(5,4)$. 
Fig.~\ref{fig:imagenette} shows the sample reconstruction of our algorithm when $f(x_1)$ needs to be recovered from the backup worker.
One can see that the recovered computation result (5th column) closely matches the missing computation result $f(x_1)$ (1st column).

\paragraph{{\ci{} with Multiple Failures}}
We first handle two failures at the same time by applying \ci{} with $(n,k) = (4,2)$.
We consider the following encoders:
{\small \begin{align*}
    \text{Enc}_1(x_1, x_2) &= f^{-1}\left(\frac{f(x_1)+f(x_2)}{2}\right),\\
    \text{Enc}_2(x_1, x_2) &= f^{-1}\left(\frac{f(x_1)+2f(x_2)}{3}\right).
\end{align*}}
With these encoders, the four tasks can be viewed as four linearly independent sums of $f(x_1)$ and $f(x_2)$, \emph{i.e.}, 
{\small \begin{align*}
\begin{bmatrix}
f(x_1)\\
f(x_2)\\
\frac{f(x_1) + f(x_2)}{2}\\
\frac{f(x_1) + 2f(x_2)}{3}
\end{bmatrix}
= 
\begin{bmatrix}
1 & 0\\
0 & 1\\
\sfrac{1}{2} & \sfrac{1}{2}\\
\sfrac{1}{3} & \sfrac{2}{3}
\end{bmatrix}
\begin{bmatrix}
f(x_1)\\
f(x_2)
\end{bmatrix}.\label{eq:1}
\end{align*}}
One can confirm that \emph{any} two rows of the coefficient matrix in the RHS is an invertible matrix.
That is, as long as any two of the four computation results are available, the decoder can recover the computation results $f(x_1)$ and $f(x_2)$. 
We evaluate this coding scheme on the \dm{} dataset. 
We consider a setting where both $f(x_1)$ and $f(x_2)$ are unavailable, \emph{i.e.}, one must recover them only from the backup task results.  
Even for this very non-trivial setting, we obtain 67.4\% degraded accuracy.
Similarly, we also test our framework with $(n,k) = (6,4)$.
Focusing on the setting where two of the first four results are not available (i.e., both the encoders are involved when decoding), we achieve $45.4\%$ degraded accuracy.
To the best of our knowledge, all the previous works focused on the $n=k+1$ setting, and this is the first non-trivial degraded accuracy for $n>k+1$.

\small
\subsection{Encoding/decoding Overhead}\label{sec:exp_encoder_overhead}
\begin{figure}[t]
    \centering
    \includegraphics[width=0.48\textwidth]{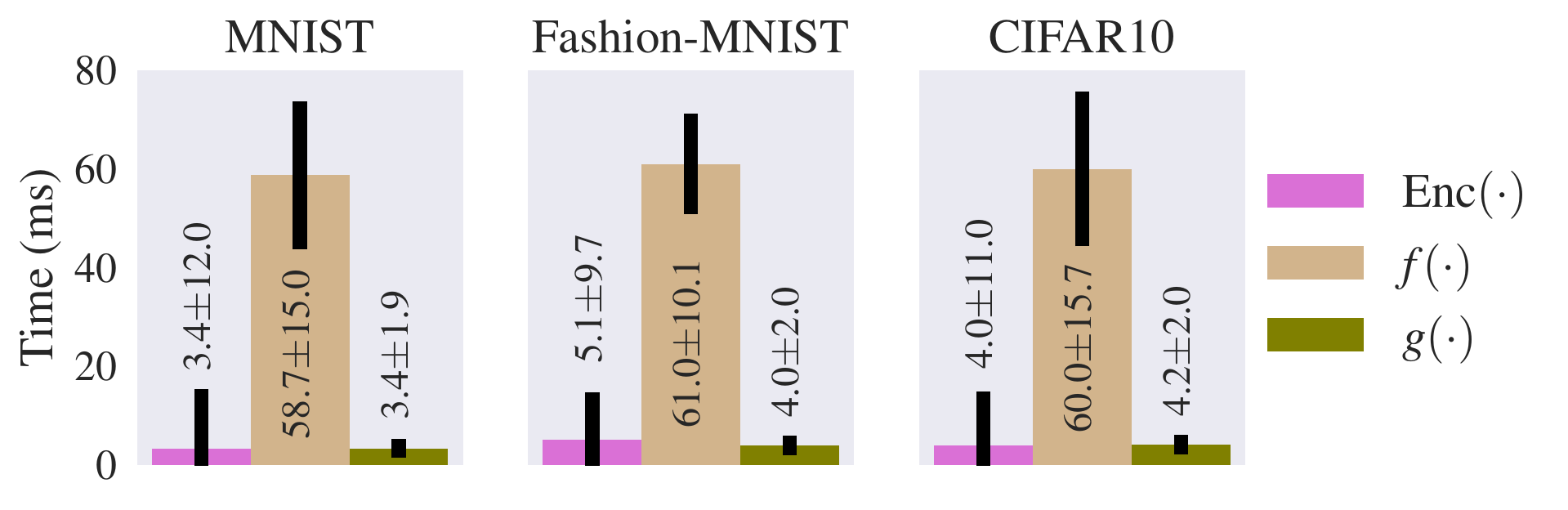}
    \vspace{-0.35in}
    \caption{\textbf{Wall-clock compute time of $\enc{(\cdot)}$, $f(\cdot)$, and $g(\cdot)$.} 
    \textit{The compute overhead of encoding must be negligible compared to the actual inference time. 
    We measure the wall-clock time to run $\enc{(\cdot)}$,  $f(\cdot)$, and $g(\cdot)$ to quantify the compute overhead of encoding when $k=2$.
    We fix the batch size as $256$ images and measure the compute time over $10$ independent runs, reporting averages, and standard deviations. 
    We can observe low overheads: $5.8\%$ for \dm{}, $8.3\%$ for \df{}, and $6.6\%$ for \dc{}.}
    }
    \label{fig:time}
    \vspace{-0.05in}
\end{figure}
\normalsize
At the time of query arrival, \ci{} must first compute $n-k$ encoded queries before assigning compute tasks to workers. 
If the encoding procedure requires a non-negligible amount of compute time, then the whole premise of coded computation is invalidated.

To avoid this, as described in Sec.~\ref{sec:exp_architectures}, we carefully chose the encoder architecture so that its compute time is negligible compared to that of the inference function. 
Shown in Fig.~\ref{fig:time} are the wall-clock compute time of $\enc{(\cdot)}$, $f(\cdot)$ and $g(\cdot)$. 
We measure the running time over $10$ runs and report the mean and standard deviation.
All of these are measured on a 12-GB NVIDIA TITAN Xp GPU, 128-GB of DRAM, and 40 Intel Xeon E5-2660 CPUs. 
Note that the time to compute $f(\cdot)$ is dominant, and the encoding time is relatively negligible.
The encoding overhead is $5.8\%$ for \dm{}, $8.3\%$ for \df{}, and $6.6\%$ for \dc{}. 

\begin{figure}[t]
    \vspace{-0.1em}
    \centering
    \includegraphics[width=0.45\textwidth]{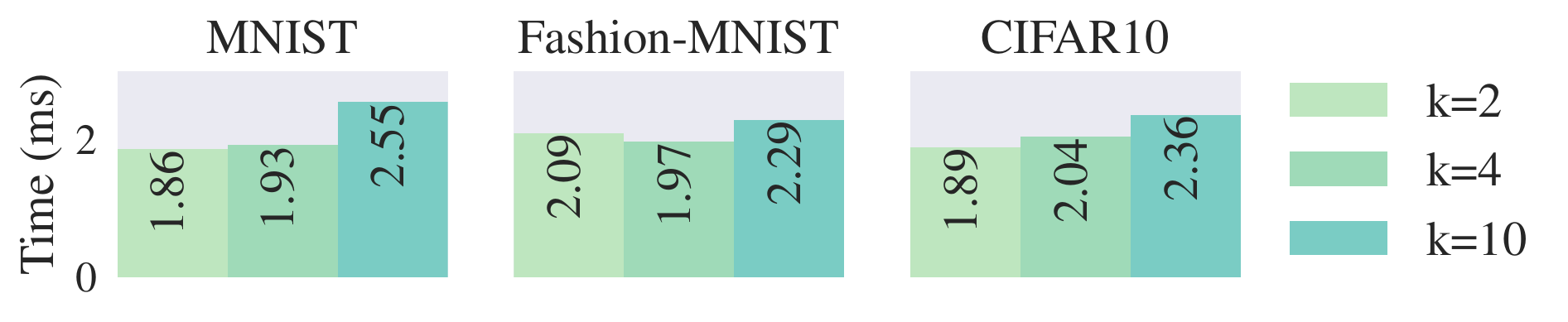}
    \vspace{-0.2in}
    \caption{\textbf{Encoding overhead as a function of $k$.} 
    \textit{The encoding overhead is measured for $k=2, 4, 10$.
    We fix the batch size as $256$ images and measure the compute time of $\enc{(\cdot)}$ over $10$ independent runs, reporting the medians. 
    The encoding overhead remains nearly constant as $k$ increases on all datasets.}
    }
    \label{fig:time2}
    \vspace{-.15in}
\end{figure}
As $k$ increases, the encoder complexity also increases.
To avoid linearly increasing complexity, in Sec.~\ref{sec:exp_architectures}, we designed our encoder architecture such that only the first few layers' complexity depends on $k$. 
Shown in Fig.~\ref{fig:time2} is the encoding overhead as a function of $k$. 
Observe that the encoding overhead remains almost constant as $k$ increases. 

The decoding overhead of \ci{} is minimal, and a simple online decoding algorithm can make it independent of $k$. 
See the supplemental materials for more details.

\textit{Take-away: \ci{}'s encoding overhead is negligible ($<10\%$) due to its lightweight encoder architecture and remains almost constant as $k$ increases. The decoding overhead does not scale with $k$ with online decoding.}

\subsection{End-to-End Latency}
\begin{figure}[t]
    \vspace{-0.5em}
    \centering
    \includegraphics[width=0.35\textwidth]{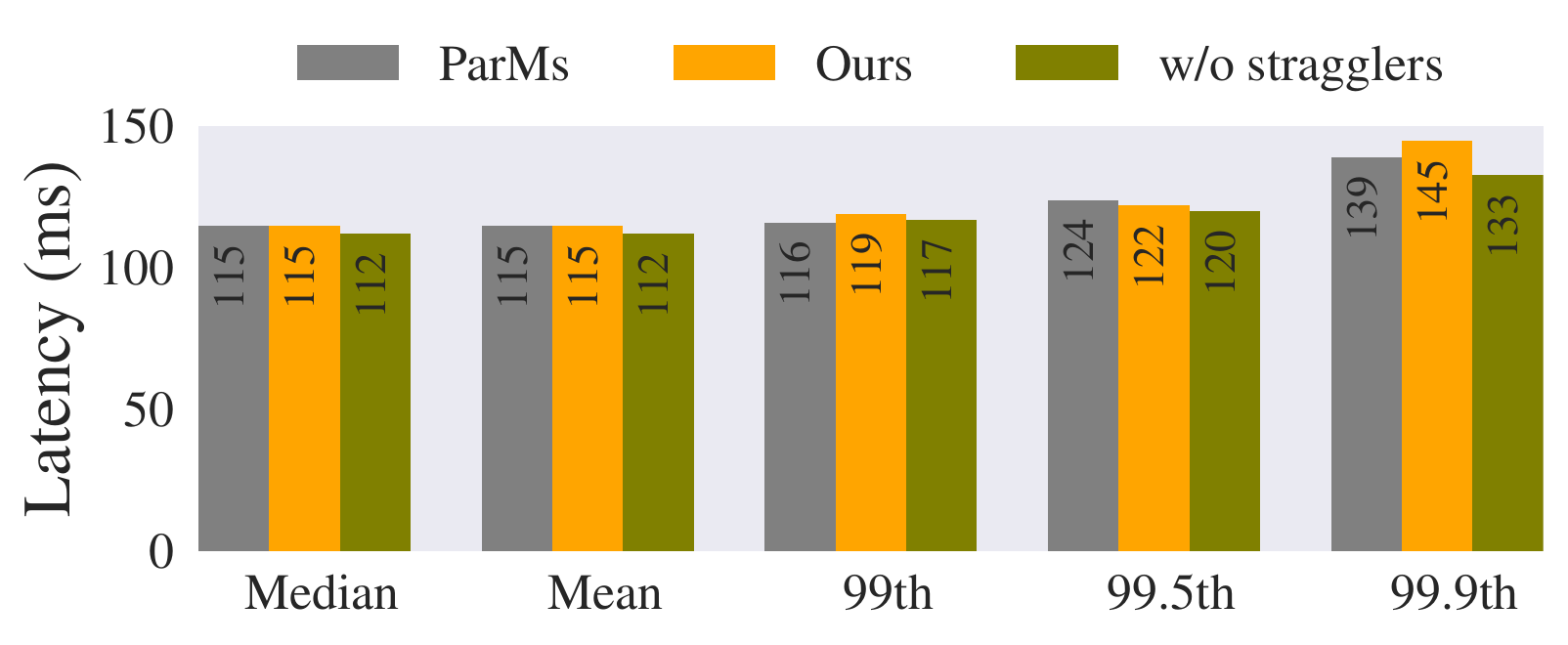}
    \vspace{-0.15in}
    \caption{\textbf{End-to-end latency on AWS cluster with $(n, k)=(11,10)$}. 
    \textit{
    Latencies are measured on \parm{} (grey), our \ci{} (orange) and inference models  without stragglers (olive). 
    \ci{} shows negligible overhead compared to \parm{} and inference models.
    For instance, the $99^{th}$-tail latency of \ci{} is only $3\si{\milli\second}$ higher than \parm{}, accounting for less than $3\%$ inference latency.
    Note that, our extra latencies (and \parm{}s) probably come from the communication to the redundancy worker.
    }}
    \label{fig:latency}
    \vspace{-0.2in}
\end{figure}

We evaluate the effectiveness of \ci{} in reducing tail latency by evaluating its end-to-end inference latency measured on an AWS EC2 cluster. 
In particular, we implement a custom testbed for \ci{} and \parm{}, written with MPI4py~\citep{dalcin2011parallel}.
All experiments are run on an Amazon AWS EC2 cluster with GPU instances (p2.xlarge with NVIDIA K80).
We deploy a pretrained inference model in $k=10$ instances and deploy the encoder and the inference model in one additional instance.
We also deploy one dedicated instance as the front-end node.
To emulate stragglers in our testbed, we randomly add artificial latencies of $0.1\si{\second}$ to one of the $k$ workers, following the setup proposed in \citep{tandon2017gradient}.
We then measure the end-to-end latency that is the time between when the front-end server receives a query and when the prediction result becomes available at the front-end server.
We measure latencies over 5,000 independent \dc{} classification queries.
\parm{} is set to use the default linear encoder/decoder and the same neural network architecture with ours.
Shown in Fig.~\ref{fig:latency} are the end-to-end latency statistics of \ci{} and \parm{}, compared with the latency statistics without artificial delays. 
As expected, \ci{}'s additional overhead is almost negligible, compared to the other latency sources such as communication and inference. 

\textbf{{Varying Batch Size}} For completeness, we evaluate end-to-end latency of \ci{} with batched queries.
Using $(n,k)=(3,2)$, with the batch size of 2 and 4, we also observe close gaps of 2.29\% and 3.78\% between \ci{} and \parm{} on $99.9$th percentile latency, respectively.

\subsection{Applicability to Multi-task Serving}\label{sec:exp_mtl} 
To illustrate the multi-tasking advantage, we perform two 2-task experiments on \dc{}: (1) fine and coarse classifications, (2) classification, and density estimation. 
The overall architecture remains the same except that we have $g_1(\cdot), g_2(\cdot)$ for two tasks instead of only one $g(\cdot)$.
In the first experiment, fine classification is with the 10 original classes, and we define two coarse classes as `vehicle' (airplane, automobile, ship, and truck) and 'animal' (the other classes).
\ci{} achieves high accuracies for both the tasks, across different $k$ (shown in Table~\ref{table:multitask} in \app{}).
For instance, with $k=4$, we achieve $86.6\%, 43.4\%$ on coarse and fine classification, respectively. 
For the second experiment, we co-train a classifier and a normalizing flow model~\citep{behrmann2018invertible} based on \ir{}, obtaining 85.2\% normal accuracy and 3.9 bits/dim, respectively.
Applying \ci{}, with $(n, k)=(5,4)$, the two models recover 50\% degraded accuracy and 6.8 degraded bits/dim.
These elementary results demonstrate the applicability of \ci{} to multi-task settings.

\subsection{{Ablation Study}}

\paragraph{Beyond Invertibility} %
Extending our framework to non-invertible architectures is a crucial future direction.
~\citep{engstrom2019adversarial} shows that a robust representation learned by adversarial training is approximately invertible.
When $f()$ is not invertible, one can still apply our framework by finding an approximate inverse of $y$ by solving $\min_x \|f(x) - y\|$. 
We replace the INN in \ci{} by the robust classifier~\citep{engstrom2019adversarial} and use approximate inverse for inversion.
On \dc{} with $(n,k)=(5,4)$, we achieve 38.5\% in degraded accuracy, which shows the promise of this direction.

\paragraph{Effect of \mmix{}}
The regularizer makes $g(\cdot)$ robust to small approximation errors.
We found that degraded accuracies on \dc{} when $k=2,4$ decrease by 42\% and 64\% when trained without \mmix{}, despite gaining higher normal accuracy.
As discussed in Sec.~\ref{sec:mixup}, this can be accounted to the effect of \mmix{} on learning smooth data manifold.
This effect increases the chance that the aggregated embedding locates on the manifold, and thus the corresponding inverses obtain the in-domain structural and semantic representation.
This makes encoder training easier as most domain translation algorithms are designed for close-to-enough domains.

\section{Conclusion}
We present \ci{}, a new coded computation-based approach for designing a resilient prediction serving system.
Inspired by a new coded computation algorithm for invertible functions, it jointly trains the inference function and the encoder, by making use of recent developments in the deep learning literature such as \mmix{} and domain translation algorithms. 
Maintaining a clear separation between encoding and inference, \ci{} provides a hassle-free design alternative to the existing methods for system designers. 
\ci{} is shown to significantly outperform the state of the art especially when the compute resource overhead is as low as $10\%$.

\section*{Acknowledgements}
This material is based upon work supported by NSF/Intel Partnership on Machine Learning for Wireless Networking Program under Grant No. CNS-2003129.

\newpage
\appendix
In (A), we present additional experiment results, including a concrete example of the illustration in Section 1.1 (A.1), additional results of degraded accuracy for different $k$ values and architectures (A.2), full results of multi-task classification (A.3), and more end-to-end latency evaluations (A.4).

In (B), we present the detail of architectures, choices of loss, training parameters, and observations.
We also present a learning curve of encoder training.

Furthermore, we provide a more detailed discussion on decoding overhead and online decoding in (C).

\section{Additional Experiment Results}

\subsection{Linear Functions on Synthesis Dataset}

To complete the story of illustration in Section 1.1, we synthesize a 2D-dataset with a rotation function.
Here, we use the setting $n=k+1$ with $k$ inputs and $n$ parallel workers. 
The inference function $f_\theta$ is the rotation function with angle $\theta = \frac{\pi}{3}$, which has rotation matrix as
$\begin{bmatrix}
   \cos\theta & -\sin \theta \\
   \sin\theta & \cos \theta
\end{bmatrix}$. 
The inverse function of $f$ has the rotation matrix as 
$\begin{bmatrix}
   \cos\theta & \sin \theta \\
   -\sin\theta & \cos \theta
\end{bmatrix}$.
Input distribution $\mathcal{D}$ is the mixture Gaussian $\frac{1}{2}\mathcal{N}(\mu_1, \Sigma) + \frac{1}{2}\mathcal{N}(\mu_2, \Sigma)$, where $\mu_1 = \begin{bmatrix} 1 \\ 0 \end{bmatrix}, \mu_2 = \begin{bmatrix} 0 \\ 1 \end{bmatrix}, \Sigma = \begin{bmatrix} 1 & 0 \\ 0 & 1\end{bmatrix}$.

For evaluation, we randomly draw a set of $k$ inputs $\{x_j\}_1^k$ from the input distribution $\mathcal{D}$.
We randomly select an input $x_a$ as the missing target to recover and remove it from the input set.
We recover $x_a$ from the $f$ values of the remaining inputs $\{x_j\}_{j\neq a}$, as
$$\hat{x_a} = kf(\frac{\sum_{j\neq a}x_j}{k}) - \sum_{j\neq a}f(x_j)$$
We measure the reconstruction error as $\|x_a - \hat{x_a}\|_2$.
We repeat this process 50K for each value of $k\in[2, 100]$, and calculate the mean of reconstruction errors.

We found that the reconstruction errors are almost negligible for all $k$ values (from $10^{-15}$ to $10^{-14}$).
These errors are probably caused by floating-point errors in computing.
This result shows that our framework exactly recovers the missing inputs.

\subsection{Additional Degraded Accuracy on More Values of $k$, and An Additional Invertible Architecture.}
\begin{figure}[t]
    \centering
    \includegraphics[width=0.48\textwidth]{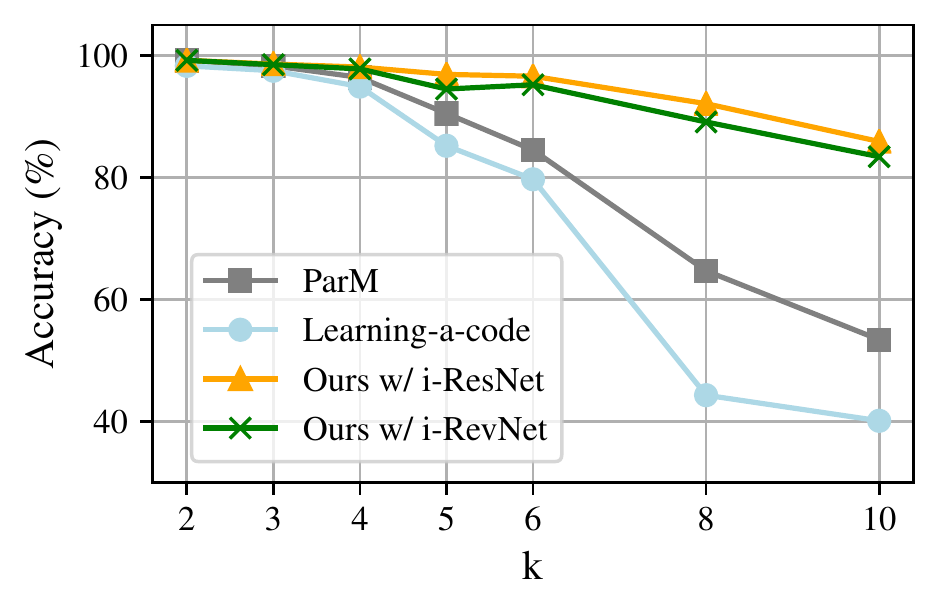}
    \vspace{-.3in}
    \caption{\textbf{Degraded Accuracy measured on \dm{}.}
    \textit{Comparison the accuracy of two baselines, \parm{} (grey), \lac{}(blue), with \ci{} built on top of \ir{} (orange) and i-RevNet (green).
    Our \ci{} models outperforms both baselines on every $k$, especially when $k$ is large.
    }}
    \label{fig:accuracy-mnist-all}
    \vspace{-0.1in}
\end{figure}

Fig.~\ref{fig:accuracy-mnist-all} presents the degraded accuracy measured on \dm{} with $k=2, 3, 4, 6, 8, 10$.
We add another invertible architecture, i-RevNet~\citep{jacobsen2018revnet} as an alternate to \ir{} in our \ci{} framework.
The result confirms our findings that \ci{} outperforms the baselines, and the gap becomes larger as $k$ increases.

\begin{table}[t]
\caption{\textbf{Illustration of \ci{} on multi-task learning.}
\textit{Normal accuracy (Normal) and degraded mode accuracy ($k=2,4,10$) on 10-class \textsc{FINE} image classification and 2-super-class \textsc{COARSE} image classification.
The two tasks use the same embedding learned by \ci{}.
While maintaining the high degraded mode accuracy on the \textsc{FINE} task, 
\ci{} achieves high accuracy on \textsc{COARSE} tasks with a modest overhead of computing (just adding the linear classifiers). 
}
}
\label{table:multitask}
\begin{center}
\begin{tabular}{|l|c|c|c|c|}
\hline
{Task} & {Normal} & {$k=2$} & {$k=4$} & {$k=10$} \\
\hline
Fine    & 86.2\%& 74.7\%& 43.4\% & 31.2\%\\
\hline
Coarse & 98.2\%&  93.5\%& 86.6\% & 71.4\%\\
\hline
\end{tabular}
\end{center}
\vspace{-0.1in}
\end{table}

\subsection{Applicability to Multi-task Serving}
Table~\ref{table:multitask} presents results of 2-task classification with $k=2,4,10$.
The two tasks are image classification on \textsc{FINE} (10-class) and \textsc{COARSE} (2-super-class) sets using the shared embedding learned by \ci{}.

We note that the normal accuracy drops in the fine classification.
Indeed, we do not need to retrain the embedding layer $f(\cdot)$ since the coarse classification task can be viewed as a sub-task of the fine classification task.
However, to mimic scenarios where we do not have such a hierarchical relationship between tasks, we retrain the embedding layer as well jointly with two task-specific classifiers $g_1(\cdot)$ and $g_2(\cdot)$, accounting for the drops in fine classification accuracy.

\subsection{Additional Results of End-to-end Latency}
\begin{figure}[t]
    \centering
    \includegraphics[width=0.35\textwidth]{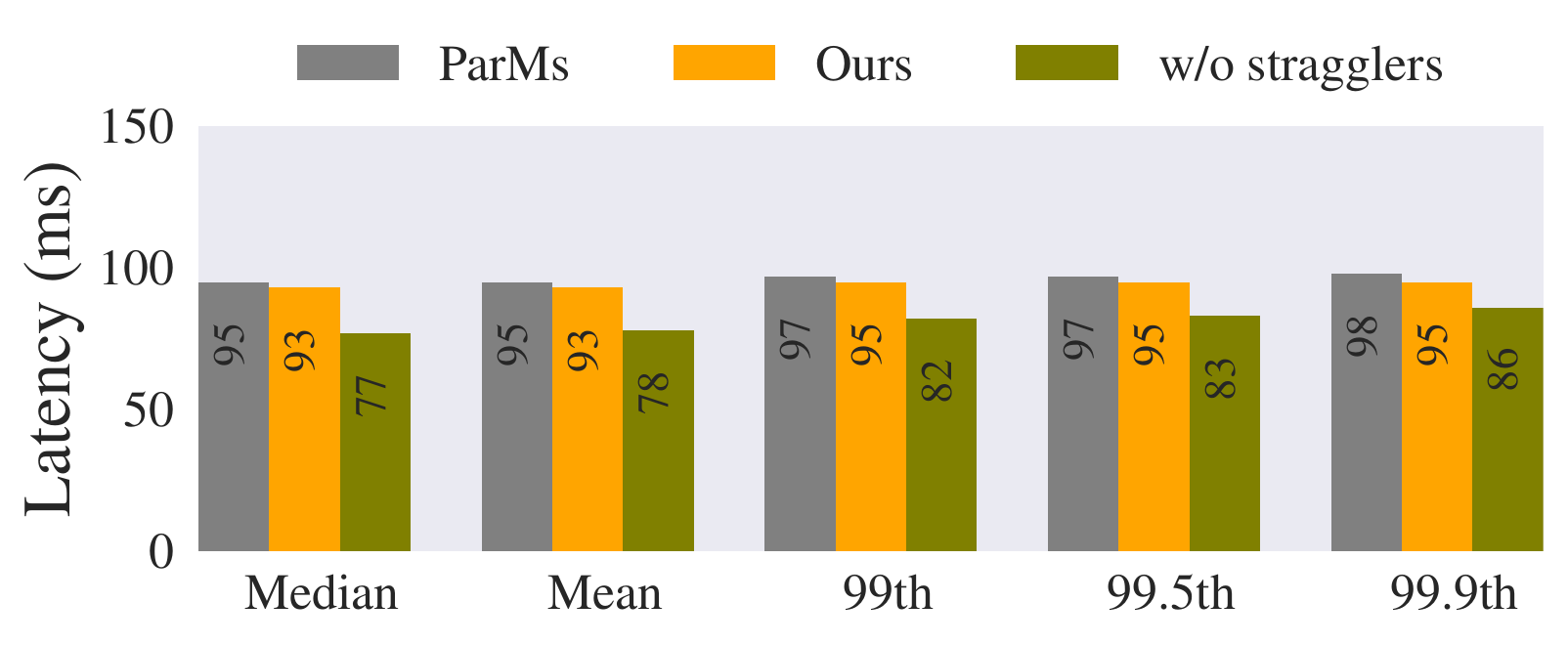}
    \includegraphics[width=0.35\textwidth]{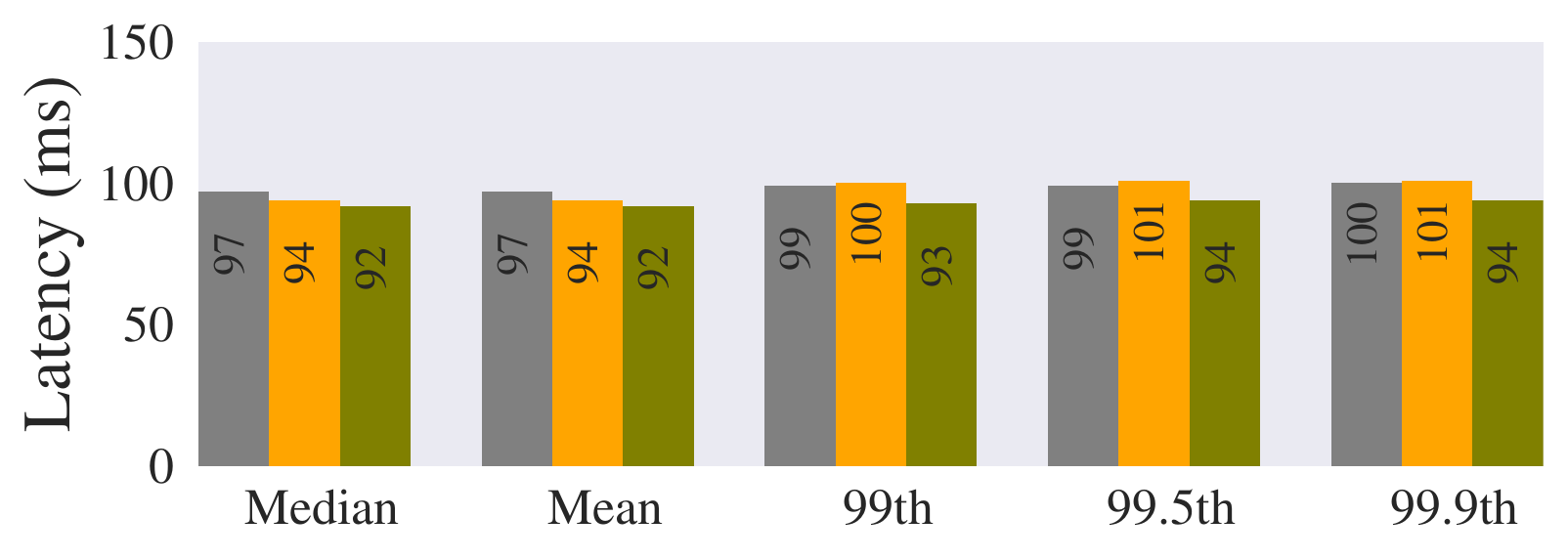}\\
    \includegraphics[width=0.35\textwidth]{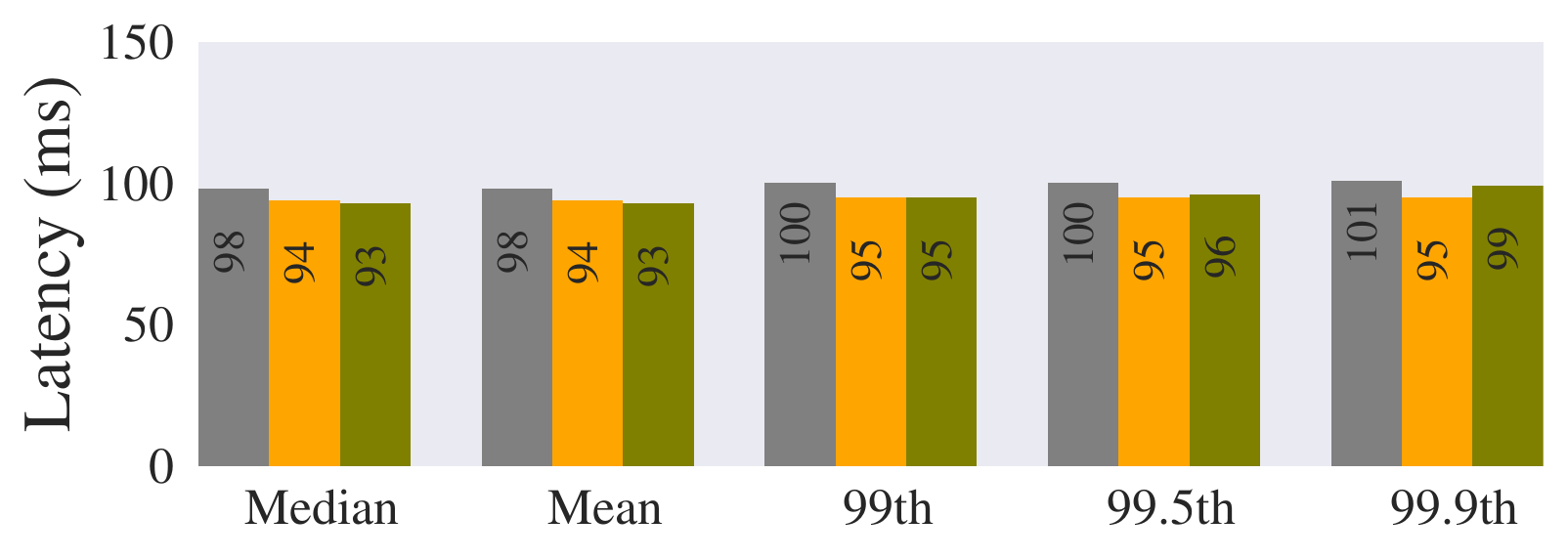}\\
    \includegraphics[width=0.35\textwidth]{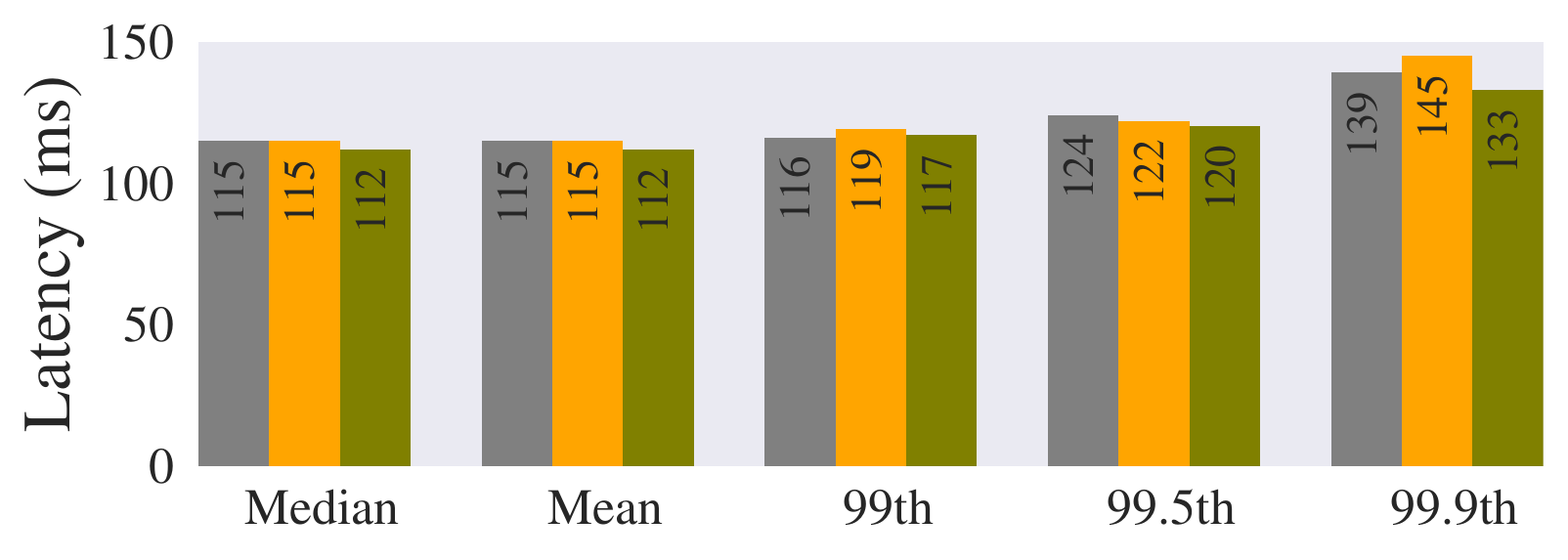}
    \caption{\textbf{End-to-end latency on AWS cluster with $k=2,3,4, 10$ (top to bottom)}. 
    \textit{
    Latencies are measured on \parm{} (grey), our \ci{} (orange) and inference models  without stragglers (olive). 
    \ci{} shows negligible overhead compared to \parm{} and inference models.
    For instance, in case $k=10$, the $99^{th}$-tail latency of \ci{} is only $3\si{\milli\second}$ higher than \parm{}, accounting for less than $3\%$ inference latency.
    Note that, compared to the setup on the inference models, we use an extra redundancy worker on  \parm{} and \ci{} ($k+2$ instances), our extra latencies (and \parm{}) probably come from the communication to the redundancy worker.
    }}
    \label{fig:latency_full}
\end{figure}

Shown in \ref{fig:latency_full} the measurements on end-to-end latency with more values of $k=2,3,4,10$, on \parm{}, \ci{}, and inference models without stragglers.
For each $k$, we use $k+2$ instances for \parm{} and \ci{} (an extra redundancy worker).
For the measurement of inference models without stragglers, we use $k+1$ instances (without redundancy).
For all values of $k$, \ci{} shows negligible overhead compared to \parm{} and inference models.

\section{Details on Architectures and Training}

\begin{figure}[t]
    \centering
    \includegraphics[width=0.4\textwidth]{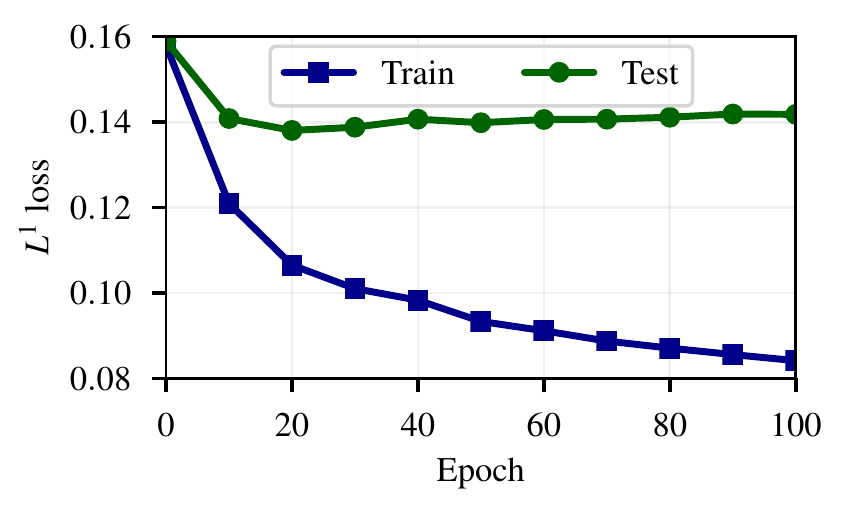}
    \vspace{-0.2in}
    \caption{\textbf{Encoder training curve.} 
    \textit{We show the encoder training curve.  Here, we train an encoder on \dm{} with $k=4$.  The loss function is a combination of GAN loss and $L_1$ loss. Observe that the $L_1$ (or $L^1$) loss on train data keeps decreasing while the test loss saturates around epoch $20$. We choose the best-performing epoch based on the validation loss.}}
    \label{fig:encoder_curves}
    \vspace{-0.2in}
\end{figure}

\paragraph{Encoding Training Curves}
Fig.~\ref{fig:encoder_curves} shows the training curve of our encoding function, with $(n,k)=(5,4)$.
We select the best encoding model based on the valuation loss.

\paragraph{\ir{} Architecture}
We use 7, 9 and 9 convolutional \ir{} blocks for \dm{}, \df{} and \dc{} respectively. 
Note that 7 and 9 \ir{} blocks correspond to ResNet-64 and ResNet-82, respectively.
For \ir{}, we remove the injective padding module that introduces zero paddings to increase spatial dimensions of images for classification performance improvement. 
This removal results in a slight classification accuracy decrease, but significantly improves the invertibility of \ir{}, especially for off-manifold embedding vectors.

\paragraph{\ptp{} Architecture}
We use the U-Net architecture for generators and the PixelGAN model~\cite{makhzani2017pixelgan} for discriminators (instead of PatchGAN in~\cite{isola2017image}), with the recommended architectures~\citep{isola2017image}.
We maintain the low encoding overhead by designing a sufficiently small architecture for the encoder.
Furthermore, one may minimize the trained encoder's inference time by applying compression techniques to the larger encoder models.

\paragraph{Choice of Loss}
We have tried different loss functions for the encoder training, including various GAN losses, regression loss, knowledge distillation loss, and their combinations.
Regression losses ($L_1$ and $L_2$ losses) do not capture well the semantic, so a small regression loss does not necessarily imply a small error in the embedding space.
In our experiments, encoder training failed when $L_1$ or $L_2$ losses were used without GAN losses, except when trained on \dm{} with $k=2$. 
Knowledge distillation (KD) loss is observed to work better than regression losses and sometimes even better than GAN loss in terms of degraded accuracies, as KD loss directly utilizes the soft labels.
However, as KD loss is specific to the classification, and it is not clear how one can use the KD loss for encoder training when different types of downstream tasks are given.
The combination of GAN loss and $L_1$ loss worked the best for most cases, but we also observed several failure cases.
When $k$ is large, the ideal encoded inputs lose most of their structural patterns and semantic representation, making GAN loss less useful. 
The design of an efficient loss function for encoder training is an interesting open problem.

\paragraph{Further Training Details}
We train the classifier with \mmix{}. 
Specifically, we apply \mmix{} on random layers (including the input layer) with the mixup coefficient ($\alpha_{\text{mixup}}$) being 1.
Each classifier is trained for 200, 400, 600 epochs on \dm{}, \df{}, and \dc{}, respectively. 
For Imagette2 dataset, we transfer the i-RevNet classifier~\citep{jacobsen2018revnet} for ImageNet to ImagetNett2 and fine-tune with \mmix{} in 100 epochs.
For \textit{the classifier training}, we use Adam optimizer~\cite{kingma2014adam} with $\beta_1 = 0.5, \beta_2 = 0.999$.
We set the learning rate as $0.1$ with 40 warming-up epochs, and decay the learning rate by $0.2$ every 60 epochs. 
The batch size is $128$.  
For \textit{the encoder training}, we train 100, 200, and 500 epochs for $k=2,4,10$ respectively, as it becomes harder to learn when $k$ increases.
We also use Adam optimizer with $\beta_1 = 0., \beta_2 = 0.9$ and lambda schedulers for both generators and discriminators. 
Learning rates are set to be $2e-4$ for both optimizers.
We train 5 iterations of the discriminator per each iteration of the generator.
The batch size is 64.
For \textit{implementation}, we use the PyTorch framework.
For the small-scale datasets, we use a single compute node consisting of a 12-GB NVIDIA TITAN Xp GPU, 128-GB of DRAM, and 40 Intel Xeon E5-2660 CPUs. 
For the large-scale dataset (ImageNette2), we use a 48-GB RTX8000 GPU.

\section{Decoding Overhead and Online Decoding}
The decoding overhead is minimal.
When $n = k+1$, the decoding procedure simply requires one scalar-vector multiplication and $k-1$ subtractions. 
To see this, recall that $\widehat{f(x_1)} := kf\left(x_{k+1}\right)-\textstyle{\sum_{i=2}^{k}}f(x_k)$.

The decoding time can be further reduced by performing online decoding.
This is possible because, in practice, not all of the $k$ tasks will complete exactly at the same time. 
Instead, their task results will be available to the decoder one by one.
Therefore, the decoder can continuously update the best-effort estimates of $\widehat{f(x_i)}$'s while receiving task results one by one.
More specifically, the decoder can run the following update algorithm at the time of task $j$ ($1\leq j \leq k+1$) completion:
\begin{align}
    \widehat{f(x_i)} = \begin{cases}
    f(x_j)&\textrm{if}~ i=j\\
    \widehat{f(x_i)}-f(x_j)&\textrm{if}~ 1\leq i\leq k, i\neq j,\\
    \widehat{f(x_i)}+kf(x_j)&\textrm{if}~ i=k+1,
    \end{cases}
\end{align}
for all $1\leq i\leq k$. 
Note that one does not have to continuously update $\widehat{f(x_i)}$ after receiving $f(x_i)$.
This online algorithm hides all the decoding overhead but one operation, minimizing the decoding overhead by a factor of $k$, i.e., the decoding overhead does not scale with $k$.

\end{document}